# Efficient Multi-Scale Attention Module with Cross-Spatial Learning

Daliang Ouyang, Su He, Guozhong Zhang, Mingzhu Luo,
Huaiyong Guo, Jian Zhan, Zhijie Huang

AEROSPACE SCIENCE & INDUSTRY SHENZHEN (GROUP) CO., LTD.,
Shenzhen, China

**Abstract**

*Remarkable effectiveness of the channel or spatial attention mechanisms for producing more discernible feature representation are illustrated in various computer vision tasks. However, modeling the cross-channel relationships with channel dimensionality reduction may bring side effect in extracting deep visual representations. In this paper, a novel efficient multi-scale attention (EMA) module is proposed. Focusing on retaining the information on per channel and decreasing the computational overhead, we reshape the partly channels into the batch dimensions and group the channel dimensions into multiple sub-features which make the spatial semantic features well-distributed inside each feature group. Specifically, apart from encoding the global information to re-calibrate the channel-wise weight in each parallel branch, the output features of the two parallel branches are further aggregated by a cross-dimension interaction for capturing pixel-level pairwise relationship. We conduct extensive ablation studies and experiments on image classification and object detection tasks with popular benchmarks (e.g., CIFAR-100, ImageNet-1k, MS COCO and VisDrone2019) for evaluating its performance. Code is available at https://github.com/YOLOonMe/EMA-attention-module.*

## 1. Introduction

Following the evolution of deep Convolutional Neural Networks (CNNs), more notable network topologies are employed in the fields of image classification and object detection tasks. It behaves the remarkable ability to enhance the learnt feature representation when we extend the CNNs to across multiple convolutional layers. However, it leads to stack more deep convolutional counterparts and needs much consumption of memory and computation resources, which is the primary drawback for constructing the deep CNNs [1], [2]. As an alternative way, the attention mechanism method, due to the flexible structure characteristics, not only strengths the learning of more discriminative feature representation, but also can be easily plugged into backbone architecture of the CNNs. Consequently, the attention mechanisms attract much interest in the research communities of computer vision.

It has been generally accepted that there are mainly three types of attention mechanisms proposed like the channel attention, the spatial attention and both of them. As the representative channel attention, Squeeze-and-excitation (SE) explicitly modeled the cross-dimension interaction for extracting the channel-wise attention [3]. Convolutional block attention module (CBAM) [4] established the cross-channel and cross-spatial information with the semantic inter-dependencies between spatial and channel dimensions in the feature maps. Consequently, CBAM shown great potential in integrating cross-dimensional attention weights into the input features. However, the manual design of the pooling operations involves complex processing that brings in some computational overhead. To overcome the shortcomings of computational cost limitations, a long-standing and effective way, using feature grouping method to divide features into multi-group on different resources, is provided [5]. Obviously, it can make each set of features well-distributed over the space. Following the setting, Spatial group-wise enhance (SGE) attention [6] grouped the channel dimensions into multiple sub-features, and improved the spatial distribution of different semantic sub-features representations, which achieves outstanding performance.

One of the most effective ways to manage model complexity is to use the convolution with channel dimensionality reduction [7]. Comparing with the SE attention, Coordinate attention (CA) [8] embedded the direction-specific information into the channel attention along spatial dimension direction, and selected an appropriate reduction ratio of channel dimensionality achieving comparable performance. On the contrary, such phenomenon is probably the most common problem that alleviates the computational burden with dimensionality reduction in the pixel-wise regression as compared with the coarse-grained CV tasks. Inspired by the thoughts that estimate the highly non-linear pixel-wise semantics, the Polarized self-attention (PSA) [9] completely collapsed the input feature maps along the counterpart channel dimensions and reserved high spectral resolution. With a small reduction ratio, PSA shown great potential in performance improvement. Although the appropriate channel reduction ratios yield better performance, it may bring side effect in extracting deep visual representations, which is explored the efficiency without dimensionality reduction in Efficient channel attention (ECA) [10].

Large layers depth plays an important role in increasing the representational ability of the CNNs. However, it inevitably leads to more sequential processing and higher latency [11], [12]. Different from the large depth attentions described as a linear sequence, Triplet attention (TA) [13] proposed a triplet parallel branches structure for capturing cross-dimension interaction against the different parallel branches. With the parallel substructures, Shuffle attention (SA) [14] grouped channel dimensions into multiple sub-features and addressed them in parallel, which can be efficiently parallelized across multiple processors. Moreover, Parallel networks (ParNet) [15] constructed the parallel sub-networks improving the efficiency of feature extraction while maintaining small depth and low latency.

Taking the inspiration from the aforementioned attention mechanisms, it can be seen that the cross-dimensional interaction contributes to the channel or spatial attention prediction. We, based on the grouping structure, revise the sequential processing method of CA and propose a novel efficient multi-scale attention (EMA) without dimensionality reduction. Note that here, only two convolutional kernels will be placed in the parallel subnetworks respectively. One of the parallel subnetworks is a 1x1 convolutional kernel that handles in the same manner as shown in CA and the other is a 3x3 convolutional kernel. To demonstrate the generality of our proposed EMA, the detailed experiments are presented in Section 4, including the results on the CIFAR-100, ImageNet-1k, COCO and VisDrone2019 benchmarks. Together with the experiment results on image classification and object detect tasks are shown in Figure.1. Our main contributions are concluded as follows:

- We propose a novel cross-spatial learning method and design a multi-scale parallel subnetworks for establishing both short and long-range dependencies.

- We consider a generic method that reshapes the partly channel dimensions into the batch dimensions to avoid some form of dimensionality reduction via a universal convolution.

- Apart from building the local cross-channel interaction in each parallel subnetwork without channel dimensionality reduction, we also fuse the output feature maps of the two parallel subnetworks by a cross-spatial learning method.

- Comparing with CBAM, Normalization-based Attention Module (NAM) [16], SA, ECA and CA, EMA not only achieves the better results, but also is more efficient in terms of required parameters.

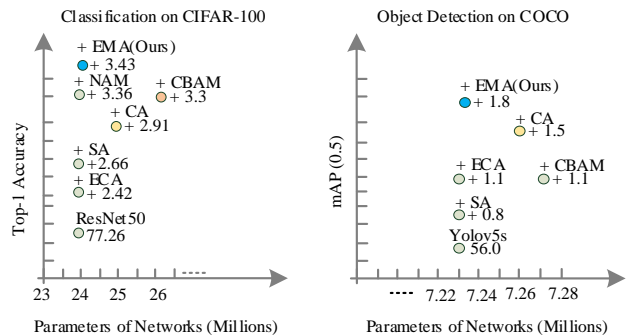

**Fig.1.** Comparing the accuracy of different attention methods with ResNet50 [17] as backbones, it shows EMA behaves the highest Top-1 accuracy while having less model complexity. Models are also evaluated on the COCO benchmark with the standard backbone of yolov5s (v6.0) [18], which illustrates that EMA is efficient yet effective.

## 2. Related Work

**Feature grouping**. Feature grouping has been studied extensively in the previous literature. To alleviate the restriction of computer computing budgets, AlexNet illustrated the grouped convolution is favorable to distribute the model over two groups on more GPU resources [19]. With increasing the number of feature grouping, ResNeXt has been proved that the activation of the sub-features will be spatially affected by different patterns and noisy backgrounds [20]. Res2Net was assumed as a hierarchical mode to transfer the grouped sub-features enabling CNNs for representing features at multiple scales [21]. Focusing on the feature grouping structure, SGE exploited the overall information of the entire group space to both strengthen the feature learning in semantic regions and compress the noise, but it failed in modeling the correlation between spatial and channel attention. To emphasize meaningful representation power of considering both channel and spatial attention features, SA divided the channel dimensions into multiple groups, and introduced those groups into two parallel branches with an equalitarian distribution method. Hence, the two branches can model the correlation between spatial and channel attention information separately. However, only part of the channels will be taken into account to exploit the inter-relationship of channels and construct informative features by fusing both spatial and channel-wise information.

**Multi-stream networks**. Given the sense that stacking one layer after another increase depth of the network for learning increasingly abstract features. As a fine-grained attention mechanism, the parallel structure was utilized by PSA to model the long-range dependencies towards high-quality pixel-wise regression task and yielded remarkable gains. Although the parallel substructures strengthen the capacity of visual representation, they bring a number of additional parameters and calculations, which is less suitable for applications. Correspondingly, the triplet attention blended cross-channel and spatial information with rotation operation into the three parallel branches for learning increasingly abstract features. However, the captured attention weights are directly aggregated by simple averaging, which is unfavorable to boost the discriminability of deep features.

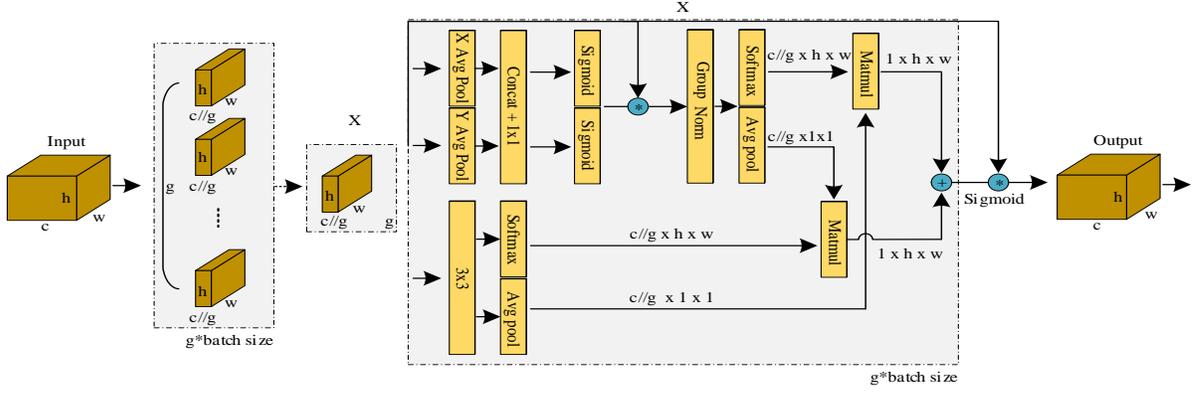

**Fig.2.** Illustration of our proposed EMA. Here, "g" means the divided groups, "X Avg Pool" represents the 1D horizontal global pooling and "Y Avg Pool" indicates the 1D vertical global pooling, respectively.

**Multi-scale convolution.** Different kernel sizes can enable the CNNs to collect multi-scale spatial information in the same processing stage. To enrich the feature space, Inception [22] presented multi-branch structure, where the local receptive fields in each branch are not fixed. Thus, the aggregation approach enables the CNNs to aggregate multi-scale information from different branches. Selective kernel networks [23] adopted an adaptive selection strategy that realizes adaptive receptive field size of neurons to effectively enriching feature representations. In addition, EPSANet [24] replaced the 3x3 convolution with the established multi-scale pyramid structure, which models a cross-channel interaction in a local manner and learns the multi-scale spatial information independently.

Comparing with the aforementioned attention modules, our proposed multi-scale attention module shows more better performance improvements. Different to the above attention methods, where the learnt attention weights are aggregated by a simple averaging method, we fuse the learnt attention maps of the parallel subnetworks by a cross-spatial learning method. It uses the matrix dot-product operations aiming at capturing pixel-level pairwise relationship and highlighting global context for all pixels [25], [26].

## 3. Efficient Multi-Scale Attention

In this section, we first revisit the coordinate attention block, where the positional information is embedded into the channel attention maps for blending cross-channel and spatial information. We will develop and analyze our proposed EMA module, in which the parallel subnetworks block helps effectively capture the cross-dimension interaction and establish the inter-dimensional dependencies.

### 3.1. Revisit Coordinate Attention (CA)

As shown in Figure. 3 (a), CA block can be firstly viewed as a similar approach to the SE attention module, where the global average-pooling operation is exploited to model the cross-channel information. Generally, the channel-wise statistics can be generated by using a global average pooling, where the global spatial position information is squeezed into a channel descriptor. Subtly different to the SE, CA embedded the spatial positional information into channel attention maps for the enhancement of feature aggregation.

Note that the CA will decomposes the original input tensors into two parallel 1D feature encoding vectors for modeling the cross-channel dependencies with spatial positional information. Firstly, one of the parallel routes is directly from a 1D global average-pooling along the horizontal dimension direction and hence can be viewed as a collection of positional information along the vertical dimension direction [8]. Let the original input tensor $X \in \mathbb{R}^{C \times H \times W}$ denotes the intermediate feature map, where $C$ means the numbers of the input channels, $H$ and $W$ indicate the spatial dimensions of the input features respectively. Consequently, the 1D global average-pooling for encoding the global information along the horizontal dimension direction in $C$ at height $H$ can be denoted by

$$z_c^H(H) = \frac{1}{W} \sum_{0 \leq i \leq W} x_c(H, i) \quad (1)$$

where $x_c$ indicates the input features at $c$-th channel. With such encoding processes, CA captures the long-range dependencies at the horizontal dimension direction and preserves precise positional information at the vertical dimension direction. Similarly, the other one route is directly from a 1D global average-pooling along the horizontal dimension direction and hence can be viewed as a collection of positional information along the vertical dimension direction. The route utilizes the 1D global average-pooling along the vertical dimension direction to capture long-range interactions spatially and preserve the precise positional information along the horizontal dimension direction, strengthening the attention to the spatial region of interest. The pooling output in $C$ at width $W$ can be formulated as

$$z_c^W(W) = \frac{1}{H} \sum_{0 \leq j \leq H} x_c(j, W) \qquad (2)$$

where $x_c$ indicates the input features at $c$-th channel. In the following, the input features can encode the global feature information and assist the model in capturing global information along two spatial directions respectively, which are in the absent of convolutions. Moreover, it generates two parallel 1D feature encoding vectors, and then permutes one vector into the other vector shape before concatenating two parallel 1D feature encoding vectors across a convolutional layer. Those two parallel 1D feature encoding vectors will share a 1x1 convolutional convolution with dimensionality reduction. The 1x1 kernel is to enable model to capture local cross-channel interaction and share the similarities with channel-wise convolutions. And then, CA further factorizes the outputs of 1x1 convolution kernel into two parallel 1D feature encoding vectors and stack one 1x1 convolutional convolution followed by a non-linearity Sigmoid function in each parallel routes respectively. Finally, the learnt attention map weights of the two parallel routes will be utilized to aggregate the raw intermediate feature map as the final outputs. Therefore, CA not only preserved the precise positional information, but also effectively exploit the long-range dependencies by encoding inter-channel and spatial information.

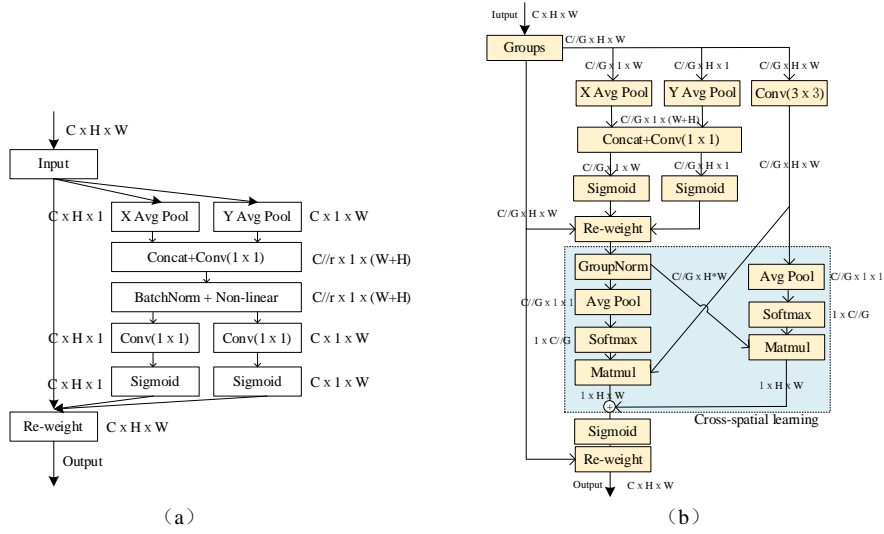

**Fig.3.** Comparisons with different attention modules: (a) CA Module; (b) EMA module.

It is self-evident that the CA embeds the precise positional information into channel-wise and captures long-range interactions spatially, achieving the impressive performance. The two 1D global average-pooling are designed for encoding the global information along two spatial dimensions direction and capture the long-range interactions spatially along different dimension directions respectively. However, it neglects the importance of the interaction among entirely spatial positions. Moreover, the limited receptive field of 1x1 kernel convolution adversely hinders for modeling of local cross-channel interaction and capitalizing on the contextual information.

### 3.2. Multi-Scale Attention (EMA) Module

The parallel substructures help the networks avoid more sequential processing and large depth. Given the above defined parallel processing strategy, we adopt it in our EMA module. The overall structure of EMA is shown in Figure. 3 (b). In this section, we will discuss how the EMA learns effective channel descriptions without channel dimensionality reduction in convolution operations, and produce a better pixel-level attention for high-level feature maps. Specifically, we only pick out the shared component of 1x1 convolution from the CA module, named it as 1x1 branch in our EMA. To aggregate multi-scale spatial structure information, a 3x3 kernel is placed in parallel with 1x1 branch for fast responses and we name it as 3x3 branch. Considering the feature grouping and multi-scale structures, it is favorable to efficiently establish both short and long-range dependency for better performance.

| Method | Backbone | #.Param. | FLOPs | Top-1 (%) | Top-5 (%) |
|---|---|---|---|---|---|
| Baseline [17] | ResNet50 | 23.71M | 1.30G | 77.26 | 93.63 |
| + CBAM [16] | ResNet50 | 26.24M | 1.31G | 80.56 | 95.34 |
| + SA | ResNet50 | 23.71M | 1.31G | 79.92 | 95.00 |
| + ECA | ResNet50 | 23.71M | 1.31G | 79.68 | 95.05 |
| + NAM [16] | ResNet50 | 23.71M | 1.31G | 80.62 | 95.28 |
| + CA | ResNet50 | 25.57M | 1.36G | 80.17 | 94.94 |
| + EMA (ours) | ResNet50 | **23.85M** | **1.32G** | **80.69** | **95.59** |
| Baseline [17] | ResNet101 | 42.70M | 2.53G | 77.78 | 94.39 |
| + CA | ResNet101 | 46.22M | 2.54G | 80.01 | 94.78 |
| + EMA (ours) | ResNet101 | **42.96M** | **2.53G** | **80.86** | **95.75** |

Table 1: Comparison of various attention methods on CIFAR-100 in terms of network parameters (in millions), FLOPs, Top-1 and Top-5 Validation Accuracies(%).

**Feature Grouping.** For any given input feature map $X \in \mathbb{R}^{C \times H \times W}$, EMA will divide $X$ into $G$ sub-features across the channel dimensions direction for learning different semantics, where the groups-style can be donated by $X = [X_0, X_i, ..., X_{G-1}], X_i \in \mathbb{R}^{C//G \times H \times W}$. Without losing generality, we let $G \ll C$ and assumed that the learnt attention weight descriptors will be utilized to strength the feature representation of interest region in each sub-feature.

**Parallel Subnetworks.** The large local receptive fields of neurons enable the neurons to collect multi-scale spatial information. Accordingly, EMA conducts that three parallel routes are exploited to extract attention weight descriptors of the grouped feature maps. Two of parallel routes is in 1x1 branch and the third one route is that the 3x3 branch. For capturing dependencies across all channels and relieving the computation budgets, we model the cross-channel information interaction at channel direction. To be more specific, there are two 1D global average pooling operations employed to encode the channel along two spatial directions respectively in 1x1 branch and only a single 3x3 kernel is stacked in 3x3 branch for capturing multi-scale feature representation.

Given the truth that there is no batch coefficient in the dimension of the convolution function for the normal convolution, the number of convolution kernels are independent of the batch coefficients of the forward operational inputs. For example, the parameter dimension of the normal 2D convolution kernel in Pytorch is $[oup, inp, k, k]$, which is not involved the batch dimensions, where $oup$ means the out planes of the inputs, $inp$ indicates the input planes of the input features and $k$ denotes the kernel size respectively. Accordingly, we reshape and permute $G$ groups into the batch dimension, and redefine the input tensor with shape of $C//G \times H \times W$. On the one hand, with similar treatment as CA, we concatenate the two encoded features against the images height direction and make it share the same 1x1 convolution without dimensionality reduction in 1x1 branch. After factorize the outputs of 1x1 convolution into two vectors, two non-linear Sigmoid functions are employed to fit the 2D Binormial distribution upon linear convolutions. For achieving different cross-channel interactive features between the two parallel routes in 1x1 branch, we aggregate the two channel-wise attention maps inside each group via a simple multiplication. On the other hand, the 3x3 branch captures the local cross-channel interaction via a 3x3 convolution to enlarge the feature space. In this way, EMA not only encodes the inter-channel information to adjust the importance of different channels, but also preserves the precise space structure information into channel.

**Cross-spatial learning.** Benefiting from the capability of building interdependencies among channels and spatial locations, there have been extensively studied and broadly used in a variety of computer vision tasks recently [27], [28]. In PSA, it exhausted the representation capacity within its channel-only and spatial-only branches, and kept the highest internal resolution in attention learning to solve the semantic segmentation. Inspired by this, we provide a cross-spatial information aggregation method at different spatial dimension direction for richer feature aggregation. Note that here, we still have introduced two tensors where one is the output of 1x1 branch and the other is the output of the 3x3 branch. Then, we utilize the 2D global average pooling to encode global spatial information in the outputs of 1x1 branch, and the outputs of the least branch will be transformed to the correspond dimension shape directly before the joint activation mechanism of channel features, i.e., $\mathbb{R}_1^{1 \times C//G} \times \mathbb{R}_3^{C//G \times HW}$ [9]. The 2D global pooling operation is formulated as

$$z_c = \frac{1}{H \times W} \sum_j^H \sum_i^W x_c(i, j) \qquad (3)$$

which is designed for encoding the global information and modeling the long-range dependencies. For efficient computation, the natural non-linear functions Softmax for 2D Gaussian maps is employed at the outputs of 2D global average pooling to fit the upon linear transformations. By multiplying the outputs of above parallel processing with matrix dot-product operations, we derived our first spatial attention map. To observe this, it collects different scale spatial information in the same processing stage. Moreover, we similarly utilize the 2D global average pooling to encode global spatial information in the 3x3 branch and the 1x1 branch will be transformed to the correspond dimension shape directly before the joint activation mechanism of channel features, i.e., $\mathbb{R}_3^{1 \times C//G} \times \mathbb{R}_1^{C//G \times HW}$.
After that, the second spatial attention map, which preserves the entire precise spatial positional information is derived. Finally, the output feature map within each group is calculated as the aggregation of the two generated spatial attention weight values followed by a Sigmoid function. It captures pixel-level pairwise relationship and highlights global context for all pixels. The final output of EMA is the same size of $X$, which is efficient yet effective to stack into modern architectures.

| Model | Datasets | #.Param. | FLOPs | mAP (0.5) | mAP (0.5:0.95) |
|---|---|---|---|---|---|
| Yolov5s [18] | COCO | 7.23M | 16.5M | 56.0 | 37.2 |
| + CBAM | | 7.27M | 16.6M | 57.1 | 37.7 |
| + SA | | 7.23M | 16.5M | 56.8 | 37.4 |
| + ECA | | 7.23M | 16.5M | 57.1 | 37.6 |
| + CA | | 7.26M | 16.50M | 57.5 | 38.1 |
| + EMA (ours) | | **7.24M** | **16.53M** | **57.8** | **38.4** |
| Yolov5x [30] | VisDrone | 90.96M | 314.2M | 49.29 | 30.0 |
| + CBAM | | 91.31M | 315.1M | 49.40 | 30.1 |
| + CA | | 91.28M | 315.2M | 49.30 | 30.1 |
| + EMA (ours) | | **91.18M** | **315.0M** | **49.70** | **30.4** |

**Table 2:** Object detection results of different attention methods on COCO and VisDrone val datasets. EMA Attention results in higher performance gain with slightly higher computational overhead.

As discussed above, we can know the attention factors are only guided by the similarities between the global and local feature descriptors inside each group. Considering the cross-spatial information aggregation method, both the long-range dependencies will be modeled, and the precise positional information are embedded into EMA. Fusing context information with different scales enables the CNNs to produce a better pixel-level attention for high-level feature maps. Subsequently, the parallelizing of the convolution kernels seems to be a more powerful structure to handle both short and long-range dependencies by using the cross-spatial learning method. In contrast to the progressive behavior of limited receptive fields formed, utilizing 3x3 and 1x1 convolutions in parallel capitalizes more contextual information among intermediate features.

## 4. Experiments

In this section, we provide the details for experiments and results to demonstrate the performance and efficiency of our proposed EMA. We conduct experiments on the challenging computer vision tasks like classification on CIFAR-100 and ImageNet-1k, and object detection on MS COCO and VisDrone2019 datasets. To verify the efficient performance, we integrate EMA into the standard network architectures like ResNet50/101 and MobileNetV2 [29], respectively. Our object detect code implementation is based upon the Pytorch YOLOv5s (v6.0) repository by Ultralytics. For image classification tasks on the CIFAR-100 and ImageNet-1k datasets, the experiments are performed with exactly the same data augmentation and training configuration settings in NAM and CA to make fair comparisons. The numbers of groups $G$ in our proposed EMA module is set as 32. All experiments run on a PC equipped with two RTX 2080Ti GPUs and on Intel(R) Xeon Silver 4112 CPU@2.60Ghz.

### 4.1. Image Classification on CIFAR-100

We investigate our proposed EMA on CIFAR-100 datasets, whose sets include the images with 32x32 pixels and consist of images drawn from 100 classes. The training set is comprised of 50k images and the validation set is comprised of 10k images. We exploit stochastic gradient descent (SGD) with momentum of 0.9 and the weight decay of 4e-5. The batch size is 128 by default. Our networks of all comparing approaches are trained for 200 epochs to make fair comparisons. After the training, we evaluate the performance of our network using the standard CIFAR Top-1 and Top-5 accuracy metrics.

As shown in Table 1, a comparison of several other attention mechanisms over the baseline of ResNet50/101 shows that integrating with EMA gains a very comparative performance with relatively small model complexity (i.e., network parameters and floating-point operations per second (FLOPs)). Comparing with the standard baseline of ResNet50, EMA achieves 3.43% gains in terms of Top-1 accuracy and 1.96% advantages over the Top-5 accuracy. With almost the same computational complexity, the Top-

1 accuracy can be improved by 0.52% by our proposed EMA as compared to the CA. In addition, using ResNet101 as the backbone model, we compare EMA with CA. Obviously, our EMA outperforms CA by a large margin with less parameters (42.96M v.s. 46.22M) and lower computational cost. It is worth noting that the gain in term of Top-1 average accuracy of CA is slightly dropped from 80.17% for the ResNet50 to 80.01% for the ResNet101 as the network architecture becoming deeper.

| Method | Backbone | #.Param. | M-Adds | Top-1 (%) | Top-5 (%) |
|---|---|---|---|---|---|
| Baseline [29] | MobileNet (v2) | 3.50 M | 300 M | 72.3 | 91.02 |
| + SE [8] | | 3.89 M | 300 M | 73.5 | - |
| + CBAM [8] | | 3.89 M | 300 M | 73.6 | - |
| + CA [8] | | 3.95 M | 310 M | 74.3 | - |
| + EMA (ours) | | **3.55M** | **306M** | **74.32** | **91.82** |

**Table 3:** Comparison with attention-based models and our EMA based on MobileNetv2 on ImageNet-1k.

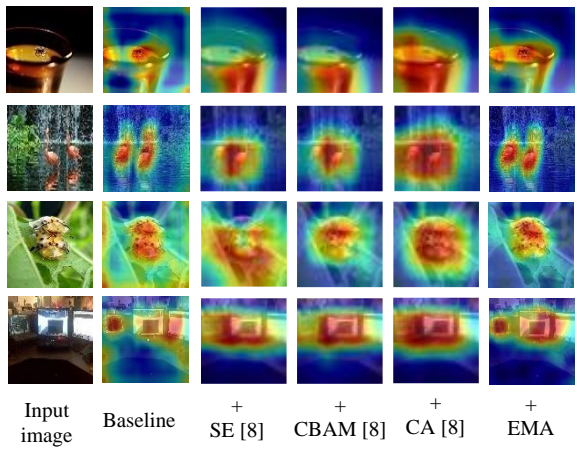

| Input image | Baseline | + SE [8] | + CBAM [8] | + CA [8] | + EMA |

Figure 4. Visualization of feature maps produced by models with different attention-based methods in the last building block of the baseline networks (MobilenetV2). It noted that our EMA module focuses on more relevant regions with more object details.

### 4.2. Image Classification on ImageNet-1k

Additionally, we compare EMA with other attention-based methods using MobileNetv2 as the baseline model on ImageNet-1k, which is a widely used large-scale benchmark for image classification. ImageNet-1K dataset consists of almost 1.28M training mages and 50K validation images with 1000 classes. During training, all images are resized to the resolution 224x224. We pick the models with best Top-1 accuracy and Top-5 accuracy performance on validation images. SGD optimizer with a learning rate of 0.4, momentum of 0.9, weight decay of 1e-5 and the linear warmup learning rate of 1e-6 are used. Following common practice, we train MobileNetv2 for 400 epochs with batch size of 256. To reduce stochastic noise during training, we use exponential moving average method.

As shown in Table 3, EMA model outperforms all other attention-based models with comparable FLOPs or multiply-adds. The baseline of MobileNetv2 model achieves 72.3% ImageNet Top-1 validation accuracy and 91.02% Top-5 validation accuracy with 300M multiply-adds. With integrating the SE module into MobileNetv2, the performance of Top-1 validation accuracy further improves to 73.5%. Furthermore, the MobileNetv2 network with the CBAM module significantly improves the Top-1 validation accuracy performance over the baseline model by 1.3%. Our EMA model achieves state-of-the-art performance of 74.4% accuracy with 306M multiply-adds, when compared to the baseline. The experimental results show that the params of the CA model is 3.95M, while the EMA model proposed in the paper is only 3.55M, which is smaller than the CA model.

### 4.3. Object Detection on MS COCO

To demonstrate the advantages of our proposed method in object detection, we investigate EMA on COCO datasets, whose training set is comprised of 118k images and the apart from the fixed size of input demanded by the original settings such as resizing the images to uniform dimensions of 640×640. The epochs and the batch size are set as 300 and 50 on one RTX 2080Ti GPUs respectively. Subsequently, we provide several attention mechanisms, 0.5 to 0.95). To further verify the effectiveness of our method, we implement all other attention strategies into the standard Yolov5s backbone with the same default settings modules into the Yolov5s backbone both improve the performance of object detection by a clear margin. Compared to CBAM, SA and ECA models, EMA and CA gaining and our EMA performs slightly better than CA in terms of mAP(0.5). In addition, it can be seen that the model size of EMA is 7.24M, which is only 0.01M lightly larger than the baseline of YOLOv5s, ECA and SA models (7.24M v.s. 7.23M). Although the FLOPs of EMA are 16.53M, which are only 0.03M larger than the baseline of YOLOv5s, EMA achieves the mAP (0.5) of 57.8% and mAP (0.5:0.95) of 38.4% on all 80 classes, which is significantly higher than other attention strategies. In general, the model size is suitable for deployment on the mobile terminals and has practical application significance.

### 4.4. Object Detection on VisDrone

Considering our proposed EMA on the dense object detection of the multi-scale feature fusion, we add a detection head for the tiny objects based on the original YOLOv5x (v6.0) [30] and integrate EMA into prediction branch to achieve the purpose of exploring the prediction potential with self-attention mechanism. During experiments, we set the size of the input image to 640×640 and we use part of pre-trained model from yolov5x for saving a lot of training time. All the attention models on VisDrone2019 trainset are trained for 300 epochs and the batch size is set as 5. The experiments are performed with exactly the same data augmentation and training configuration settings.

We use the YOLOv5x as our backbone CNN for the object detection on VisDrone datasets, where the CA, CBAM and EMA attentions are integrated into the detector respectively. As observed from the results of Table 2, both the CA, CBAM and EMA can boost the baseline performance for the object detection. We can see our proposed EMA module consistently outperforms the base CA and CBAM based networks in terms of the mAP (0.5) and mAP (0.5:0.95) respectively. It is noteworthy that CBAM boosts the performance of YOLOv5x by 0.11% and is higher than that of CA at the cost of more parameters and computations. For CA, it almost obtains the same performance as the baseline and surpass the YOLOv5x by 0.01% in terms of the mAP (0.5), while CA achieves higher parameters and computations than EMA (91.28M v.s. 91.18M and 315.2M v.s. 315.0M). Specifically, EMA adds 0.22M more parameters than baseline method, which have an improvement of 0.31% over YOLOv5x on mAP (0.5) and 0.4% on mAP (0.5:0.95) with the slightly higher parameters. These results demonstrate that EMA is an efficient module for object detection task, and further proves the effectiveness of the EMA method in this paper.

| Method | Datasets | #.Param. | FLOPs | Top-1 (%) | Top-5 (%) |
|---|---|---|---|---|---|
| + EMA_no | | 23.84M | 1.32G | 78.24 | 94.89 |
| + EMA_16 | CIFAR100 | 24.44M | 1.34G | 80.35 | 95.44 |
| + EMA_32 | | **23.84M** | **1.32G** | **80.69** | **95.59** |

**Table 4:** Ablation on relative training configuration settings on the CIFAR100.

## 5. Ablation Study

We choose ResNet50 as the baseline network and validate the importance of cross-spatial learning method by conducting ablation experiments to observe the impact of different hyperparameters in EMA, such as EMA_no (i.e., without no cross-spatial learning), EMA_16 (i.e., group size is set as 16) and EMA_32 (i.e., group size is set as 32). Comparing with EMA_32, a relatively high FLOPs and network parameters will be resulted by setting group size as 16. This is mainly due to reshape the channel dimensions into the batch dimensions that decreases the model parameters. EMA is able to call upon to distribute the model over multiple channels on more batch dimensions and process them. Moreover, we also conduct the ablation study by conducting cross-spatial learning method and the other turns off. From the view of Table 4, EMA_32 with the cross-spatial learning method outperforms EMA_no scheme. For the similarly FLOPs and network parameters, the Top-1 and Top-5 rates of EMA_32 are much higher, at 80.69% and 95.59%, respectively.

## 6. Conclusion

In this paper, we systematically investigate the properties of attention mechanisms, which leads to a principled way to combine them into CNNs. Moreover, we present new insight into how the CNNs can enjoy both good generalization and computation budgets by using a generic method that avoids some form of dimensionality reduction via a universal convolution. Due to the flexible and light-weighted characteristics, our proposed EMA can be easily exploited into different computer vision tasks for achieving best performance. We believe our EMA is more applicable to broader applications like semantic segmentation and can be stacked into other deep CNNs structure for significantly enhancing the feature representation ability. We will leave them for future work.